\documentclass{article} % One-column default

\usepackage{preprint}

\usepackage{amsmath, amsthm, amssymb, amsfonts}

\usepackage[numbers,square]{natbib}
\usepackage[utf8]{inputenc}	% allow utf-8 input
\usepackage[T1]{fontenc}	% use 8-bit T1 fonts
\usepackage{xcolor}		% colors for hyperlinks
\usepackage[colorlinks = true,
            linkcolor = purple,
            urlcolor  = blue,
            citecolor = cyan,
            anchorcolor = black]{hyperref}	% Color links to references, figures, etc.
\usepackage{booktabs} 		% professional-quality tables
\usepackage{nicefrac}		% compact symbols for 1/2, etc.
\usepackage{microtype}		% microtypography
\usepackage{lineno}		% Line numbers
\usepackage{float}			% Allows for figures within multicol
\usepackage{multicol}		% Multiple columns (Method B)

\usepackage{algorithm}
\usepackage{algpseudocode}
\usepackage{hyperref}       % hyperlinks
\usepackage{url}            % simple URL typesetting\
\usepackage{systeme}
\usepackage{nicefrac}       % compact symbols for 1/2, etc.
\usepackage{microtype}      % microtypography
\usepackage{graphicx}
\usepackage{color}
\usepackage{fancyvrb}
\usepackage{array}
\usepackage{makecell}
\usepackage{subcaption}

\usepackage{newfloat}
\DeclareFloatingEnvironment[name={Supplementary Figure}]{suppfigure}
\usepackage{sidecap}
\sidecaptionvpos{figure}{c}

\usepackage{titlesec}
\titlespacing\section{0pt}{12pt plus 3pt minus 3pt}{1pt plus 1pt minus 1pt}
\titlespacing\subsection{0pt}{10pt plus 3pt minus 3pt}{1pt plus 1pt minus 1pt}
\titlespacing\subsubsection{0pt}{8pt plus 3pt minus 3pt}{1pt plus 1pt minus 1pt}

\newcommand{\citationneeded}[1][]{\textsuperscript{[citation needed]}}
\algnewcommand{\LineComment}[1]{\State \(\triangleright\) #1}

\title{Neurogenetic Programming Framework \\ for Explainable Reinforcement Learning}

\usepackage{xwatermark}
\newwatermark[firstpage,color=gray!90,angle=0,scale=0.28, xpos=0in,ypos=-5in]{*correspondence: \texttt{v.liventsev@tue.nl}}

\usepackage{authblk}

\author[1\thanks{\tt{v.liventsev@tue.nl}}]{Vadim Liventsev}
\author[2]{Aki H{\"a}rm{\"a}}
\author[3]{Milan Petkovi{\'c}}
\affil[1,3]{Eindhoven University of Technology}
\affil[1,2,3]{Philips Research Eindhoven}

\begin{document}

\maketitle
\title{Neurogenetic Programming Framework}

\begin{abstract}
%Program synthesis is an effective way of leveraging expert knowledge in machine learning settings.
%A generative model can be pre-trained on a program written by an expert and then used to re-write this program to improve it's performance on a given benchmark.
Automatic programming, the task of generating computer programs compliant with a specification without a human developer, is usually tackled either via genetic programming methods based on mutation and recombination of programs, or via neural language models.
We propose a novel method that combines both approaches using a concept of a virtual neuro-genetic programmer \footnote{not to be confused with neuroevolution \cite{neuroevolution}: using evolutionary methods as an alternative to gradient descent for neural network training}, or scrum team.
We demonstrate its ability to provide performant and explainable solutions for various OpenAI Gym tasks, as well as inject expert knowledge into the otherwise data-driven search for solutions.
Source code is available at \url{https://github.com/vadim0x60/cibi}
\end{abstract}
\vspace{0.35cm}

\begin{multicols}{2}

\section{Introduction}\label{sec:intro}

Automatic programming is a discipline that studies application of mathematical models to infer programs from data and use these generated programs to solve various tasks.
For example, given a dataset of clinical decision-making generate a program that describes how doctors make treatment decisions based on clinical history available to them.

One of the primary motivations behind automatic programming is enabling an exchange of knowledge between human experts and machine learning models: black box models have achieved impressive results in diverse decision support settings \cite{deeprl}, sometimes competing with human experts in the field. The advantage of automatic programming systems is that they can also \emph{cooperate} with the experts:
\begin{itemize}
    \item Code generation models can be trained to produce programs similar to what experts wrote, incorporating expert knowledge into the model
    \item Code generation models can generate new programs by applying modifications to expert-written programs, using them as the basis
    \item Experts can examine the generated programs, understand the algorithm suggested by the system and learn from it
\end{itemize}

% Vadim: just some brain dumping for an intro candidate
The benefits of regulating the program induction task, especially in limited data conditions have been demonstrated earlier \cite{metainduction}. The programming language used in program induction is a regulating constraint that may be expected to guide the learning algorithm to favor solutions with operations that are most natural for the given language \cite{massiveparallelism}. For example, a strictly sequential language such as C seems most natural in cases where the expected solution is a sequential protocol. A parallel language, such as various cellular automata models or deep neural networks, favor concurrent solutions. 

We may also approach this from the angle of algorithmic information theory. The Kolmogorov complexity of the solution corresponds to the length of the program and the number of free parameters \cite{kolmogorov}. In the case of a black box deep model, the Kolmogorov complexity is an architectural constant of the parallel network model while in program induction it is optimized by the learning algorithm based on the available training data. When data is limited, the PI training may use {\sl Occam's razor  \footnote{"We consider it a good principle to explain the phenomena by the simplest hypothesis possible" \cite[book 3, chapter 2]{ptolemy}, misattributed \cite{occam} to William of Ockham }} to find a minimal-complexity solution matching the training data, while a standard NN solution always has the same complexity. 

In fact, the use of program synthesis for solving a machine learning problem can be seen as {\sl generalized hyperparameter optimization}; the program optimization may, in principle, converge to a program that implements a particular deep neural network optimized for the problem. 
% or is this philosophical bullshit going too far for your taste already ;)
%AH: added last paragraph by reprogramming later text
 In \emph{genetic programming} \cite{genprog,genprogast} new programs are generated by mutating and mixing a \emph{population} of programs. 
 A more recent approach, largely drawing on the earlier success of deep neural language models (see CodeBERT \cite{codebert} inspired by BERT \cite{bert}), have been to train black box \emph{neural models} that generate executable programs as text \cite{brain-coder,deepcoder,structural}. 
 Neural program synthesis and genetic programming both have unique advantages \cite{geneticvsneural}. 
 In this paper we propose a novel hybrid of the two families of methods. We call the method \emph{Instant Scrum} in reference to a popular Agile software team work model \cite{scrum}. We show that \emph{Instant Scrum}, IS, can solve several reinforcement-guided program synthesis tasks in standard OpenAI gym benchmarks tasks (section \ref{sec:tasks}). 

After the introduction to the relevant background in the next section, we will give a detailed description of the proposed IS methodology. Next, we describe the experimental setup and the OpenAI gym test tasks, and the results of the experiments in several variations of the core IS method. Finally, we discuss the results and propose future research directions and potential applications for hybrid neurogenetic programming. 

% TODO: Evolutionary strategies in RL

\section{Background}
% \subsection{Reinforcement Learning}

% TODO: Review solutions, not just problems

\emph{Specification} is central to the field of automatic programming: if no requirements to generated programs are specified, the task becomes random program generation - not useful in most real-world settings beyond testing compilers \cite{random1,random2}.
The field of automatic programming can be subdivided by what type of specification is used.

One way to specify expected behavior of a program is a dataset of input-output pairs \cite{deepcoder,brain-coder}.
A lot of work in this area has been inspired \cite{flashmeta,flash1,flash2} by the task of learning a formula that fits a series of cells in spreadsheet applications \cite{flashfill}.

Alternatively, the program's pre- and postcondition predicates can be described in a formal language: \emph{proof-theoretic synthesis} \cite{prooftheoretic} is the task of generating a program such that if it's input conforms to the precondition, its output has to conform to the postcondition.

In the task of \emph{semantic parsing} \cite{hearthstone,semparsing1,semparsing2}, specification is a textual description of an algorithm that has to be translated from  natural language into machine language.

Finally, specification can manifest in the form of a \emph{reinforcement learning environment} \cite{pirl} - a program defines behavior of an agent that interacts with its environments and receives positive and negative rewards from it.
The goal is to find a program that maximises rewards.
In this work, we choose RL specification, because it generalizes other methods: input-output pairs can be seen as an environment that negatively reinforces a metric of difference between expected and observed program output, while in \emph{proof-theoretic synthesis} the reward is determined by the postcondition.  
Many real-world problems, such as robot control \cite{robotrl} and clinical decision making \cite{heartpole,gym-sepsis} are formulated as reinforcement learning as well.

\subsection{Programmatically Interpretable Reinforcement Learning}
\label{sec:pirl}

More concretely, we model the task as {\em Episodic Partially Observable Markov Decision Process}:

\begin{multline}
M = (\mathcal{S}_{nt}, \mathcal{S}_t, \mathcal{A}, \mathcal{O}, p_o(o | s, a), p_s(s_\text{next} | s_\text{prev}), p_r(r | s, a), p_\text{init}(s))
\end{multline}

Here, $\mathcal{S}_{nt}$ is the set of {\em non-terminal (environment) states} and $\mathcal{S}_{t}$ is the set of {\em terminal states}. 
$\mathcal{A}$ is the set of {\em actions} that the learning agent can perform, and $\mathcal{O}$ is the set of {\em observations} about the current state that the agent can make. 

A \emph{reinforcement learning episode} starts in a state $s \in \mathcal{S}_{nt}$ sampled from $p_\text{init}$, the {\em initial distribution} over environment states.
An agent action $a \in \mathcal{A}$ at the state $s$ causes the environment state to change probabilistically, and the destination state follows the distribution $p_s(\cdot | s, a)$. 
At state $s$, the probability of making observation $o$ is $p_o(o | s)$ and the probability of obtaining \emph{reward} $r$ is $p_r(r|s,a)$. 
The process continues until a $p_s$ yields a terminal state $s \in \mathcal{S}_t$.
This distinction is what sets \emph{episodic} POMDP popularized by OpenAI gym \cite{gym} apart from the more traditional approach \cite{pomdp1,pomdp2} where the process is infinite.

A \emph{program} is a sequence of tokens

\begin{equation}
    c=(c^{(1)},c^{(2)},\dots)
\end{equation}

that defines behavior of an agent.
Depending on the programming language and implementation choices the tokens can be characters or higher-level tokens, i.e. keywords.
We denote the language's \emph{alphabet}, i.e. the set of all possible tokens as $\mathcal{L}$.

An \emph{interpreter} is a tuple $\langle \alpha,\mu \rangle$ where $\mu(c_k,m_k,o_k)$ is the \emph{memorization function} that defines how the agent's memory updates upon making an observation and $\alpha(c, m_k)$ is the \emph{action function} defines which action the agent at a certain memory state takes in the POMDP.
Memory is intialized at state $m_\text{init}$

The agent's goal is maximizing total reward collected in the environment, calculated as follows:

\begin{algorithm}[H]
\begin{algorithmic}[1]
\caption{Evaluating total reward for a program}
\Function{$\mathit{Eval}$}{$c$}
\State $R_\text{tot} \gets 0$
\State $m \gets m_\text{init}$
\State $s \sim p_\text{init}(s)$
\While{$s \in \mathcal{S}_{nt}$}
\LineComment{Observe}
\State $o \sim p_o(o | s)$
\State $m \gets \mu(c,m,o)$
\LineComment{Act}
\State $a \gets \alpha(c, m)$ 
\State $r \sim p_r(s,a)$
\LineComment{Get rewarded}
\State $R_\text{tot} \gets R_\text{tot} + r$
\LineComment{Next state}
\State $s_\text{next} \sim p_s(s_\text{next} | s, a)$
\State $s\gets s_\text{next}$
\EndWhile
\State \Return $R_\text{tot}$
\EndFunction
\end{algorithmic}
\end{algorithm}

Since the algorithm for computing this function involves repeatedly sampling values from distributions, function $\mathit{Eval}(c)$ is a mapping from the set of programs to the set of real-valued random variables.

\emph{Programmatically Interpretable Reinforcement Learning} \cite{pirl} is the task of maximizing expected total reward with respect to $c$, i.e. finding a program $c$ that is best for a given POMDP environment:

\begin{equation}
    \mathbb{E}(\mathit{Eval}(c)) \longrightarrow \max_{c}
    \label{eq:pirlgoal}
\end{equation}

\section{Methodology}
\label{sec:methodology}

How does one manage a composition of code generators in such a way that the composition yields better programs than individual contributors are capable of? 
This question is studied extensively in software project management literature \cite{mythicalmanmonth}.
And while, admittedly, project management literature is concerned with human developers and, admittedly, there exist considerable differences between human developers and mathematical models of code generation \cite{bugfixing}, we mitigate these differences with several simplifying assumptions.

\subsection{Modeling the codebase}

Following from traditional genetic programming, we define a \emph{population} of programs. 
The \emph{codebase} is a tuple of 2-tuples, representing a program $C_c^{(i)}=c$ and the total rewaard it collected $C_R^{(i)}=R_\text{tot} \sim \mathit{Eval}(c)$ (see section \ref{sec:quality}):

\begin{equation}
    \mathcal{C} = \langle \langle C_c^{(1)}, C_R^{(1)} \rangle, \langle C_c^{(2)}, C_R^{(2)} \rangle \dots \langle C_c^{(|C|)}, C_R^{(|C|)} \rangle \rangle
\end{equation}

However, unlike in traditional genetic programming, the \emph{initial population} can (optionally) be empty.

\subsection{Modeling a software developer}
\label{sec:developer}

A software developer can:
\begin{enumerate}
    \item Check out programs from the codebase $\mathcal{C}$
    \item Output new a program $c$
    \item Receive feedback on their program's quality $q$ 
    \item Learn from the feedback by mofidying its strategy
\end{enumerate}

Thus, a developer is a 2-tuple of a program distribution $p_\text{dev}(c | \theta, \mathcal{C})$ and a parameter update procedure $\mathit{Update}(\theta, c, q)$

Distribution $p_{\text{dev}}(c | \theta, \mathcal{C})$ is defined over programs and is parametrized with learnable parameters $\theta$ as well as codebase $\mathcal{C}$. 
Having codebase as parameter enables the developer to generate new programs as a modification and/or combination of existing programs, i.e. to apply genetic programming.

Learnable parameters $\theta$ encode the developer's current methodology of programming that can be modified upon receipt of positive or negative feedback using the developer's update procedure. 

The \emph{team} of developers is a tuple of 2-tuples:
\begin{equation}
    \mathcal{T} = \langle \langle \mathcal{T}_p^{(1)}, \mathcal{T}_\text{upd}^{(1)} \rangle, \langle \mathcal{T}_p^{(2)}, \mathcal{T}_\text{upd}^{(2)} \rangle \dots \langle \mathcal{T}_p^{(|\mathcal{T}|)}, \mathcal{T}_\text{upd}^{(|\mathcal{T}|)} \rangle \rangle
\end{equation}

\subsection{Modeling program quality}
\label{sec:quality}

We define two empirical metrics of overall program fitness.
The first is \emph{empirical total reward}:

\begin{equation}
    R(c|\mathcal{C}) = \frac{\sum\limits_{i=1}^{|C|} \mathbb{I}[C_c^{(i)}=c] C_R^{(i)}}{\sum\limits_{i=1}^{|C|} \mathbb{I}[C_c^{(i)}=c]}
\end{equation}

If a program has been tested in the environment ($\textit{Eval()}$ function) several times, there will be several copies of it in the codebase with different quality samples.
Averaging over them yields an unbiased estimate of the expectation from equation \ref{eq:pirlgoal}, $\mathbb{E}(\mathit{Eval}(c))$, that we set out to maximize.

The second is \emph{empirical program quality}, defined as

\begin{equation}
    Q(c|\mathcal{C}) = \frac{\sum\limits_{i=1}^{|C|} \mathbb{I}[C_c^{(i)}=c] e^{C_R^{(i)}}}{\sum\limits_{i=1}^{|C|} \mathbb{I}[C_c^{(i)}=c]}
    \label{eq:empquality}
\end{equation}

\emph{Empirical program quality} is an unbiased estimate of $\mathbb{E}(e^{\mathit{Eval}(c)})$
The idea behind exponentiating the total reward is to encourage \emph{exploration} \cite{exploration}.
Programs that on average perform poorly, but sometimes, stochastically, collect high rewards, will have a higher $Q(c|C)$ than $R(c|C)$.
We consider these programs to be \emph{high-quality additions to the codebase} because they contain the knowledge necessary for solving environment $M$, even if on average they don't solve it.
We hypothesize that applying \emph{genetic operators} (section \ref{sec:genetic}) to programs with high $Q(c|C)$ can yield programs with high $R(c|C)$.
For this reason we train developers to maximize $Q$, but when the training is complete, we pick programs from the codebase with the highest $R$ as "best programs".
 
$Q(c|C)$ has an additional technical advantage over $R(c|C)$: invariant $Q(c|C) \geq 0$ holds for all $c$.
This lets one sample programs from the codebase with probabilities proportional to their quality, see eq. \ref{eq:genmixture}.

\subsection{Populating the codebase}

Just like \emph{instant run-off voting} achieves similar results to \emph{exhaustive ballot runoff voting}, but does it much faster by replacing a series of ballots cast in a series of elections with a ballot cast once that goes on to participate in a series of virtual elections \cite{votingsystems}, our \emph{Instant Scrum} algorithm does the same to Scrum \cite{scrum}: it simulates the iterative software development process recommended by Scrum methodology without humans in the loop making it possible to run many sprints per second:

\begin{algorithm}[H]
\begin{algorithmic}[1]
\caption{Instant Scrum with a team of developers}
\label{alg:instantscrum}
\Procedure{InstantScrum}{$T,\mathcal{C},N_\text{max}$}
\State $N \gets 0$
\While{$N < N_\text{max}$}
\LineComment{For each developer in the team}
\For{$i = 1,2,\dots,|T|$}
\LineComment{Sample a program from the developer}
\State $c_\text{new}\sim p_i(c | \theta_i, \mathcal{C})$ 
\State $R_\text{tot} \gets \mathit{Eval}(c_\text{new})$
\Comment{Test the program}
\LineComment{Save the code and test result to the codebase}
\State $\mathcal{C} \gets \mathcal{C} \cup \{\langle c_\text{new}, R_\text{tot} \rangle\}$
\State $\mathit{Update}_i(\theta_i, c, q)$
\Comment{Train the developer}
\State $N \gets N+1$
\Comment{Increment sprint counter}
\EndFor
\EndWhile
\EndProcedure
\end{algorithmic}
\end{algorithm}

To combine genetic programming and neural program synthesis we introduce 3 types of developers: \emph{genetic} and \emph{neural} and \emph{dummy}, create a team that contains developers of all types and run \emph{Instant Scrum}.

\subsection{Genetic developers}
\label{sec:genetic}

A genetic developer writes programs by
\begin{enumerate}
    \item Selecting one of the 7 available stochastic \emph{genetic operators} (described below)
    \item Selecting two programs from the \emph{codebase} $C$ (parents $c_1$ and $c_2$) 
    \item Using the operator to modify the parents and yield a new (child) program
\end{enumerate}

A genetic operator is a probability distribution $p_\text{op}$ over child programs given 2 parent programs. 

\begin{equation}
    p_\text{op}(c_\text{child}|c_1,c_2)
\end{equation}

Operators whose $p_\text{op}$ is invariant to $c_2$ and depends on $c_1$ only are called \emph{mutation} operators: they generate a new program by \emph{mutating} one program $c_1$.
The rest are called \emph{combination} operators as they \emph{combine} 2 existing programs to generate a new one.

\end{multicols}

\begin{figure}
    \centering
    \begin{subfigure}{0.4\linewidth}
    \centering
    \includegraphics[width=0.4\linewidth]{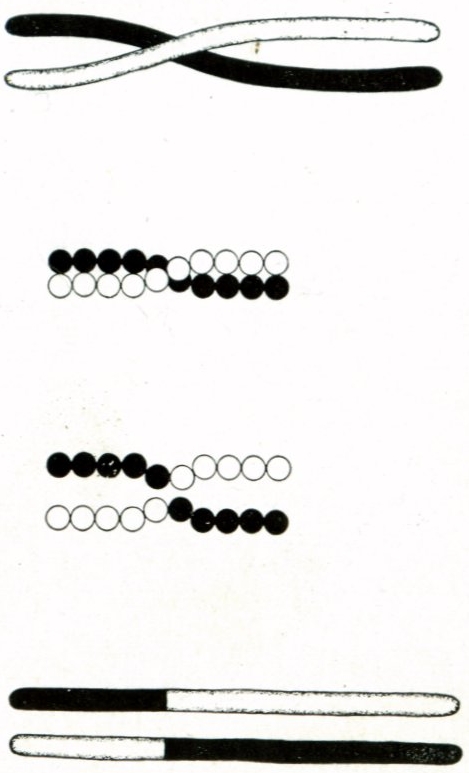}
    \includegraphics[width=0.45\linewidth]{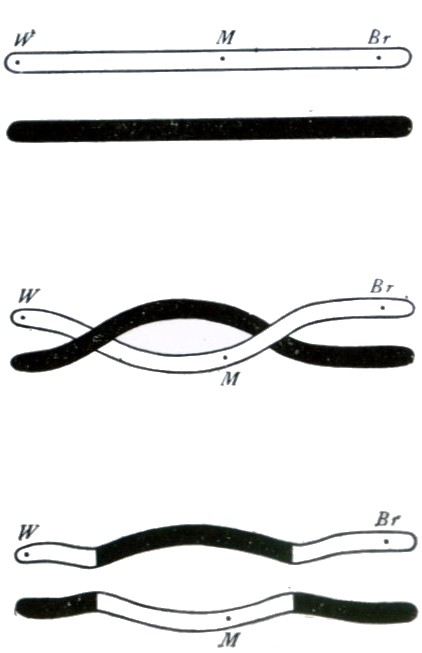}
    \caption{1-point and 2-point crossover \cite{evocritique}}
    \label{fig:crossover}
    \end{subfigure}
    \begin{subfigure}{0.4\linewidth}
    \begin{tabular}{r|l}
         Parent 1 & \color{blue}\verb|ae>>>>>34+| \\
         Parent 2 & \color{red}\verb|a[e>-a-]b[e>>-b-]| \\
         \midrule
         Shuffle mutation & \color{blue}\verb|>>4+>3>e>a| \\
         Uniform mutation & \color{blue}\verb|ae|\color{black}\verb|@|\color{blue}\verb|>|\color{black}\verb|!|\color{blue}\verb|>>3|\color{black}\verb|5|\color{blue}\verb|+| \\
         1-point crossover & \color{blue}\verb|ae>>>>|\color{red}\verb|-]b[e>>-b-]| \\
         2-point crossover & \color{blue}\verb|ae>|\color{red}\verb|>-a-|\color{blue}\verb|34+| \\
         Uniform crossover & \color{blue}\verb|ae|\color{red}\verb|e|\color{blue}\verb|>|\color{red}\verb|-|\color{blue}\verb|>>3|\color{red}\verb|b|\color{blue}\verb|+| \\
         Messy crossover & \color{blue}\verb|ae>>>>|\color{red}\verb|e>-a-]b[e>>-b-]| \\
         Pruning & \color{blue}\verb|e>>>>>4+| \\
    \end{tabular}
    \caption{All operators applied a pair of BF++ \cite{bf++} programs}
    \end{subfigure}
    
    \caption{Genetic operators by example}
    \label{fig:genops}
\end{figure}

\subsubsection{Mutation operators}

The simplest method for randomly modifying a program is \emph{shuffle mutation}: randomly re-order the tokens of $c_1$.
Let $\mathrm{A}$ be the set of all possible permutations of size $|c_1|$. $|\mathrm{A}|=|c_1|!$. 
Then

\begin{equation}
    p_\text{shuffle}(c_\text{child}|c_1,c_2) =
            \frac{\sum\limits_{\alpha \in \mathrm{A}} \mathbb{I}[\alpha(c_\text{parent}) = c_\text{child}]}{|c_1|!}
\end{equation}

Another approach is \emph{uniform mutation} where a loaded coin is tossed for every token in $c_1$. 
With probability $p_\text{ind}$ it is replaced with a random token from the alphabet $\mathcal{L}$ of the programming language, with probability $1-p_\text{ind}$ it stays the same.
The evolution of a single token under shuffle mutation is defined by distribution

\begin{equation}
    p(c^\text{new} | c^\text{old}) = \frac{p_\text{ind}}{|L|} +  (1 - p_\text{ind}) \mathbb{I}[c^\text{new} = c^\text{old}]
\end{equation}

Hence over full programs the operator is defined as

\begin{equation}
    p_\text{unimut}(c_\text{child}|c_1,c_2) = \mathbb{I}[|c_\text{child}|=|c_1|] \\ 
    \prod\limits_{i=0}^{|c_1|}  \left(\frac{p_\text{ind}}{|L|} +  (1 - p_\text{ind}) \mathbb{I}[c_\text{child}^{(i)} = c_1^{(i)}] \right)
\end{equation}

\subsubsection{Combination operators}

The combination operators we propose are all variants of \emph{crossover} - a classic genetic programming technique rooted in the way a pair of DNA molecules exchanges genes during mitosis and meiosis, displayed on figure \ref{fig:crossover}.

In DNA \cite{evocritique}, as well as in most genetic programming literature \cite{genprog,genprogast} the crossover operator combines 2 parent sequences to produce 2 children.
In this section, in order to reduce complexity, we define the distributions as if only the first child program is saved and the second one is forgotten.
Since program pair $\langle c_2, c_1 \rangle$ is equally likely to be selected for combination as $\langle c_1, c_2 \rangle$ (see eq. \ref{eq:genmixture}) this modification does not affect the resulting genetic developer distribution.

In \emph{one-point crossover} a random cut position $k$ is selected and the trailing sections of 2 parent programs beginning with the cut point are swapped with each other. 
If the parent programs have different lengths, the cut point has to fit within both programs:

\begin{equation}
    |c_1,c_2| = \min\{|c_1|, |c_2|\}
\end{equation}

\begin{equation}
    2 \leq k \leq |c_1,c_2|
\end{equation}

Hence the probability of $c_\text{child}$ being born out of \emph{one-point crossover} is

\begin{equation}
    p_\text{1ptcx}(c_\text{child}|c_1,c_2) =
        \frac{\mathbb{I}[|c_\text{child}|=|c_2|]}{|c_1,c_2|-1}
        \sum\limits_{k=2}^{|c_1,c_2|} \prod\limits_{i=1}^{k-1} \mathbb{I}[c_\text{child}^{(i)} = c_1^{(i)}] \prod\limits_{i=k}^{|c_2|} \mathbb{I}[c_\text{child}^{(i)} = c_2^{(i)}]
\end{equation}

\emph{Two-point crossover} is similar, but instead of swapping the trailing ends of programs, a section in the middle of the programs is chosen, determined by randomly selected cut-off indices $k_1$ and $k_2$ and swapped:

\begin{equation}
    p_\text{2ptcx}(c_\text{child}|c_1,c_2) =
        \frac{2 \mathbb{I}[|c_\text{child}|=|c_1|]}{(|c_1,c_2|-2)(|c_1,c_2|-1)}
        \sum\limits_{k_1=2}^{|c_1,c_2|-1}
        \sum\limits_{k_2=k_1+1}^{|c_1,c_2|} \\
        \prod\limits_{i=1}^{k_1-1} \mathbb{I}[c_\text{child}^{(i)} = c_1^{(i)}] \prod\limits_{i=k_1}^{k_2-1} \mathbb{I}[c_\text{child}^{(i)} = c_2^{(i)}]
        \prod\limits_{i=k_2}^{|c_1|} \mathbb{I}[c_\text{child}^{(i)} = c_1^{(i)}]
\end{equation}

\emph{Uniform crossover} mirrors \emph{uniform mutation} in that a loaded coin is tossed for each token in $c_1$. With probability $p_\text{ind}$ the token is replaced, but the replacement is not drawn randomly from the alphabet. Instead, the replacement comes from $c_2$:

\begin{equation}
    p_\text{unicx}(c_\text{child}|c_1,c_2) = \prod\limits_{i=0}^{|c_1|} \mathbb{I}[|c_\text{child}|=|c_1|] \\ \left(p_\text{ind} \mathbb{I}[c_\text{child}^{(i)} = c_2^{(i)}] + (1 - p_\text{ind}) \mathbb{I}[c_\text{child}^{(i)} = c_1^{(i)}] \right)
\end{equation}

Finally, \emph{messy crossover} is a version of \emph{one-point crossover} without the assumption that both parent programs have to be cut at the same index $k$.
In \emph{messy crossover}, one parent is cut at index $k_1$, another is cut at index $k_2$ and the head of one is attached to the tail of the other:

\begin{equation}
    p_\text{messy}(c_\text{child}|c_1,c_2) =
        \frac{1}{(|c_1,c_2|-1)^2}
        \sum\limits_{k_1=2}^{|c_1,c_2|} \sum\limits_{k_2=2}^{|c_1,c_2|} \\ \mathbb{I}[|c_\text{child}|=k_1+|c_2|-k_2] \\ \prod\limits_{i=1}^{k_1-1} \mathbb{I}[c_\text{child}^{(i)} = c_1^{(i)}] \prod\limits_{i=1}^{|c_2|-k_2} \mathbb{I}[c_\text{child}^{(k_1+i)} = c_2^{(k_2+i)}]
\end{equation}

\begin{multicols}{2}
\subsubsection{Pruning operator}

After initial experiments  we found that generated programs often contain sections of unreachable code or code that makes changes to the execution state and fully reverses them.
To address this, we introduced an additional operator for removing dead code (\emph{pruning}): when Instant Scrum encounters a successful program, pruning helps separate sections of this program that led to its success from sections that appeared in a highly-rated program by accident.  

Implementation of the pruning operator depends on the programming language at hand, here we define it as a pruning function $c_\text{pruned}=\mathit{Prune}(c_1)$ that outputs a program functionally equivalent to $c_1$ (memory functions $(\alpha,\mu)$ of $c_\text{pruned}$ are equal to that of $c_1$) and $|c_\text{pruned}| \leq c_1$ and a degenerate probability distribution:

\begin{equation}
    p_\text{prune}(c_\text{child}|c_1,c_2)= \begin{cases}
        1 & c_\text{child} = \mathit{Prune}(c_1) \\
        0 & \text{otherwise}
        \end{cases}
\end{equation}

\subsubsection{Operator and parent selection}
\label{sec:selection}

Let $\mathcal{P}_\text{genetic}$ be a tuple of all available genetic operators, in order of introduction, i.e. $\mathcal{P}_\text{genetic}^{(1)}=p_\text{shuffle}$ and $\mathcal{P}_\text{genetic}^{(4)}=p_\text{2ptcx}$

Genetic developer's program distribution is a mixture distribution, combining different operators that can be applied, weighted by learnable parameters, and different programs that can be sampled from the codebase, weighted by \emph{empirical quality} (eq. \ref{eq:empquality}).

\begin{equation}
    p_\text{genetic}(c | \theta, \mathcal{C}) = 
    \sum\limits_{c_1}^{C}  
    \sum\limits_{c_2}^{C} 
    \frac{Q(c_1|C) Q(c_2|C)}{(\sum\limits_{c}^{C} Q(c|C))^2} 
    \sum\limits_{i=0}^{|\mathcal{P}_\text{genetic}|} 
    \theta_i \mathcal{P}_\text{genetic}^{(i)} (c|c_1,c_2)
    \label{eq:genmixture}
\end{equation}

This is a true probability distribution if and only if $\sum\limits_{i=0}^{|\mathcal{P}_\text{genetic}|} 
    \theta_i = 1$

\subsubsection{Training the genetic developer}

One challenge that remains to be solved to fully define the genetic developer (folowing section \ref{sec:developer}) is to define a learning from feedback strategy $\mathit{Update}_\text{genetic}$.
To do this, we notice that equation \ref{eq:genmixture} contains a \emph{multi-armed bandit} \cite{banditproblem} hiding in plain sight.
Indeed, once the \emph{genetic} developer samples $c_1$ and $c_2$ from the codebase, it has to pick one of 7 available options (pull one of 7 \emph{levers}) to then receive a reward $\mathit{Eval}(c_\text{child})$.
This subproblem can be represented with a POMDP of its own and solved using one of the standard bandit algorithms \cite{banditsolutions}.

Following \emph{Occam's razor}, we picked the simplest method, \emph{epsilon-greedy optimization}: we calculate the value of each operator as mean total reward of programs generated with this operator:

\begin{equation}
    V^{(i)} = \frac{1}{|\mathcal{C}(\mathcal{P}_\text{genetic}^{(i)})|} 
    \sum\limits_{k=1}^{|\mathcal{C}(\mathcal{P}_\text{genetic}^{(i)})|}
    \mathcal{C}(\mathcal{P}_\text{genetic}^{(i)})_R^{(k)} 
\end{equation}

where $C(\mathcal{P}_\text{genetic}^{(i)})$ is the subset of the codebase produced via operator $\mathcal{P}_\text{genetic}^{(i)}$.

The $\mathit{Update}_\text{genetic}$ procedure recalculates values $V$ and sets operator probabilities to

\begin{equation}
    \theta_i = \frac{\epsilon}{|\mathcal{P}_\text{genetic}^{(i)}|} +
    \mathbb{I}[i = \underset{i}{\arg\max} V^{(i)}] (1 - \epsilon)
\end{equation}

where $\epsilon$ is a hyperparameter responsible for regulating the \emph{exploration-exploitation tradeoff} \cite{banditsexplo}

In future work, however, other bandit optimization algorithms can be used in its place \footnote{Our open-source software implementation allows for drop-in replacement of bandit algorithms}.

\subsubsection{Hyperparameters}
\label{sec:genhyper}

The genetic developer, as described above, has 2 hyperparameters:

\begin{enumerate}
    \item $p_\text{ind}$ defines severity of mutation in $p_\text{unimut}$ and $p_\text{unicx}$
    \item $\epsilon$ defines learnability of genetic operator distribution
\end{enumerate}

Note that the \emph{team} mechanism afforded by \emph{Instant Scrum} can be used not only to combine genetic and neural program synthesis, but also to combine several genetic developers with different hyperparameters.

\subsection{Neural developers}
\label{sec:neural}

The \emph{neural developer}, also known as the \emph{senior developer} because of their unique ability to write original programs, is an LSTM \cite{lstm} network followed by a linear layer that generates a sequence of vectors $h_{1},h_{2},h_{3},\dots$ where $h_i \in \mathbb{R}^{|\mathcal{L}| + 1} \forall i$ and $j$-th element of vector $h_i$, $h_i^{(j)}$, represents the probability of $i$-th token of the program being $j$-th token in the alphabet, $p(c^{(i)}=\mathcal{L}^{(j)})$.
The last element of the vector represents a special \emph{end of program} symbol.
This vector depends deterministically on the full set of neural network parameters (LSTM and linear layer) $\theta$ and can be represented as a function $h_i(\theta)$.
Then

\begin{equation}
    p_\text{neural}(c | \theta, \mathcal{C}) = h_{(|c|+1)}^{\mathcal{L}+1}
    \prod\limits_{i=1}^{|c|}
    \sum\limits_{j=1}^{|\mathcal{L}|} \mathbb{I}[c^{(i)}=\mathcal{L}^{(j)}]
    h_i(\theta)
\end{equation}

For the $\mathit{Update}_\text{neural}$ procedure we use the algorithm proposed in \cite{brain-coder}.
The subproblem of generating a program $c$ is considered as a reinforcement learning episode of it's own, where tokens are actions and token number $|c|+1$ (\emph{end of program} token) is assigned reward $q = e^R; R \sim Eval(c)$. 
In this subenvironment $h_i(\theta)$ is the policy network \cite[chapter 13]{thebook} trained using REINFORCE algorithm with Priority Queue Training.
This algorithm involves a priority queue of best known programs: we implement it as programs from $C$ with highest $Q(c|C)$ which means that the neural developer can train on programs written by other developers.

$h_i(\theta)$ can also represent several LSTM layers stacked or a different type of recurrent neural network, i.e. GRU \cite{gru}.
Hyperparameters of this neural network, such as hidden state size and/or number of stacked layers are hyperparameters of the neural developer.  

\subsection{Dummy developer}

The last developer we introduce is the simplest one:

\begin{equation}
    p_\text{dummy}(c_\text{child}|c_1,c_2) = 
    \frac{Q(c_\text{child}|C)}{\sum\limits_{c}^{C} Q(c|C)} 
    \label{eq:dummy}
\end{equation}

Dummy developer does not generate novel programs.
Instead, it uses the same quality-weighted program sampling as in equation \ref{eq:genmixture} to decide which existing program to copy.
Their utility may not be obvious at first, but note (section \ref{sec:quality}) that when the same program is added to the codebase several times, it's total reward and quality estimates are averaged and grow more accurate.

Dummy developer is a smart compromise between speed at which \emph{Instant Scrum} (algorithm \ref{alg:instantscrum}) is searching the program space and the quality of it's working map of the program space, focusing on its most "interesting" (high $Q(c|C)$) parts. 
Without dummy developer, all empirical total rewards $E[Eval(c)]$ would be low quality estimates of true fitness of the program and one spurious success of an otherwise bad program could steer the search in the wrong direction.
On the other hand, we could test each program many times before adding it to the codebase, but that would slow down the search prohibitively. 

\section{Experimental setup}
\label{sec:experiments}

\subsection{Teams}

In the table below, we introduce 5 teams.
Neural developers are denoted as lstm(hidden state dimensionality), several numbers mean a stacked LSTM.
Genetic developers are denoted as $\text{gen}(p_\text{ind},\epsilon)$, see secion \ref{sec:genhyper}.
$T_\text{small}$ and $T_\text{large}$ are recommended configurations while $T_\text{genetic}$, $T_\text{neural}$ are \emph{ablation studies} to prove that combination of neural and genetic methods is useful.

\begin{tabular}{r|c|c|c|c}
     Developer & $T_\text{small}$ & $T_\text{large}$ & $T_\text{genetic}$ & $T_\text{neural}$  \\
     $\text{lstm}(10)$ & & \checkmark & & \\
     $\text{lstm}(50)$ & & \checkmark & & \\
     $\text{lstm}(256)$ & & \checkmark & & \\
     $\text{lstm}(10,10)$ & & \checkmark & & \\
     $\text{lstm}(50,50)$ & \checkmark & \checkmark & & \checkmark \\
     $\text{lstm}(256,256)$ & & \checkmark & & \\
     $\text{gen}(0.2,0.2)$ & \checkmark & &  & \\
     $\text{gen}(\frac{1}{3},0.2)$ & & \checkmark & & \\
     $\text{gen}(\frac{1}{6},0.2)$ & & \checkmark & & \\
     $\text{gen}(\frac{1}{12},0.2)$ & & \checkmark & & \\
     dummy & \checkmark & \checkmark & \checkmark & \checkmark \\
\end{tabular}

\subsection{Language}

Instant Scrum can be used to generate programs in any programming language provided:
\begin{enumerate}
    \item An interpreter $\langle \alpha,\mu \rangle$, see section \ref{sec:pirl}
    \item A known finite alphabet $\mathcal{L}$
    \item A pruning function $\mathit{Prune}(c)$
\end{enumerate}

%Moreover, pruning function and finite alphabet are, in a sense, optional requirements: pruning isn't essential to the method, so a dummy pruning function $P(c)=c$ can be used instead and if a language's alphabet is infinite, it is usually possible to develop a reasonable finite subset of the full alphabet and generate programs in it.

% Ode to BF++

% A word on trees
The complexity of the chosen language is important since in complex languages random perturbations of program source code often produce grammatically invalid programs.
This issue has been addressed with \emph{structural models} \cite{grammargp,structural} \cite[chapter 4]{genprogast}, however, we sidestep the issue entirely by using \emph{BF++} \cite{bf++} - a simple language developed for \emph{programmatically interpretable reinforcement learning} where most random combinations of characters are valid programs.
Each BF++ command is represented with a single character, thus the only way to tokenize it is to let tokens $c^{(1)},c^{(2)},c^{(3)},\dots$ be single characters.

\subsection{Tasks}
\label{sec:tasks}

\begin{figure}[t]
    \centering
    \subfloat[CartPole-v1]{
        \centering
        \includegraphics[height=1.3in]{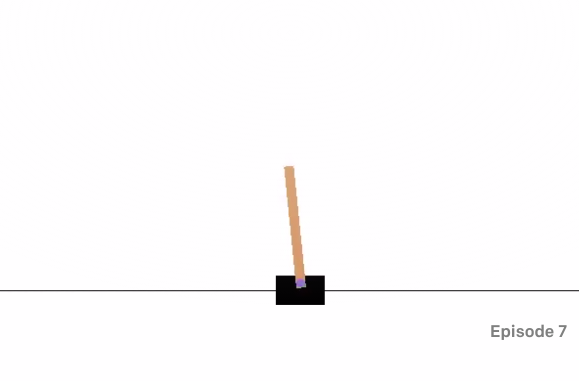}
    }
    \subfloat[MountainCarContinuous-v0]{
        \centering
        \includegraphics[height=1.3in]{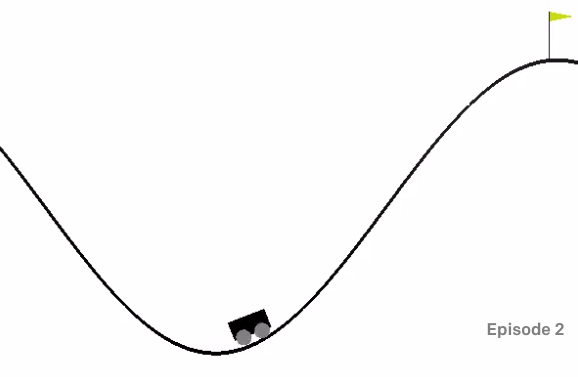}
    }
    \subfloat[Taxi-v3]{
        \centering
        \includegraphics[height=1.3in]{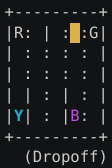}
    }
    \subfloat[BipedalWalker-v2]{
        \centering
        \includegraphics[height=1.3in]{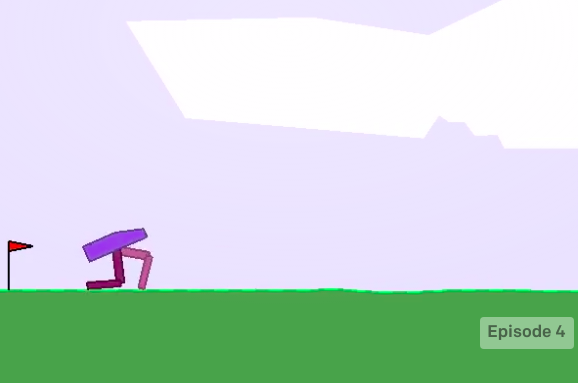}
    }
    \caption{Selected tasks, visualized}
    \label{fig:envs}
\end{figure}

Following from \cite{bf++} we synthesize programs for \textbf{CartPole-v1} \cite{cartpole}, \textbf{MountainCarContinuous-v0} \cite{mountain_car}, \textbf{Taxi-v3} \cite{taxi} \textbf{BipedalWalker-v2}  OpenAI Gym \cite{gym} environments, see figure \ref{fig:envs}.

\subsection{Initial populations}

Where possible, we run all experiments twice - a control experiment with empty intial codebase, and an experiment where codebase is pre-populated with human-written programs from \cite{bf++}.
Exceptions to this rule are 
\begin{itemize}
    \item Teams $T_\text{genetic}$ and $T_\text{pure}$ that only have code modification (not generation) capability and thus require initialization 
    \item \textbf{BipedalWalker-v2} environment, because no programs for this environment were provided in \cite{bf++}
\end{itemize}

\subsection{Stopping and scoring}

For Taxi we set an $N_\text{max}$ to $100000 |T|$ sprints, meaning every developer in the team trains for 100000 iterations.
For other tasks we used Exponential Variance Elimination \cite{evestop} early stopping algorithm to stop the process when the positive trend in $Eval(c)$ is not present for 10000 sprints.
This approach rules out the hypothesis that \emph{Instant Scrum} is equivalent to enumerative search and it finds good programs by exhaustion as opposed to learning - if that was the case, early stopping would fire immediately.
Taxi environment is treated differently because programs that cannot pick up and drop off at least one passenger are always rewarded with -200 and at first it takes many iterations to synthesize at least one program that can.
In addition to these stopping rules, a hard timelimit was set.

After the process is stopped, we pick 100 programs with the highest $R(c|C)$ and make sure each of them has been tested at least 100 times, otherwise we run $Eval(c)$ and add result to the codebase until 100 samples is reached. 

\subsection{Implementation}

We implemented the framework with Python and Tensorflow as well as DEAP \cite{deap} for genetic operators.
It is available at \url{https://github.com/vadim0x60/cibi}

\section{Results}
\label{sec:results}

See table \ref{tab:results} for a summary of best programs generated.
The metric used, average $R$ over 100 evaluations is the same metric that's used in the OpenAI gym leaderboard, so we include the threshold required to join the leaderboard for context.
\emph{Initial programs} refers to the best program in the codebase before \emph{Instant Scrum} starts when it is prepopulated with  programs from \cite{bf++}.

The main hypothesis of this paper is \textbf{confirmed}: neurogenetic approach is superior to neural program induction or genetic programming separately.
Besides, one unintuitive result of our experiments is that initialization of the codebase with previously available programs can be harmful, see $T_\text{large}$.
Overall, best results were acheived without inspiration from human experts, however, it is very valuable for lightweight teams with few small (in terms of $|\theta|$) developers.

Additionally, we can explore $T_\text{large}$ to see which of its many developers actually produced the best programs:

\begin{tabular}{r|c|c|l}
    Task & Init & $R(c|C)$ & Developer \\
    \midrule
    CartPole-v1 & & 157.35 & lstm(256) \\
CartPole-v1 & \checkmark & 57.47 & lstm(256,256) \\
MountainCar & & 91.65 & lstm(10,10) and pruning \\
MountainCar & \checkmark & 91.42 & lstm(50) \\
Taxi-v3 & & -32.12 & lstm(50) \\
Taxi-v3 & \checkmark &  -150.44  & human \\
BipedalWalker-v2 & & 8.13 & lstm(256,256) \\
\end{tabular}

The same is true for $T_\text{small}$:

\begin{tabular}{r|c|c|l}
    Task & Init & $R(c|C)$ & Developer \\
    \midrule
    CartPole-v1 & & 60.93 & lstm(50,50)  \\
CartPole-v1 & \checkmark & 143.9 & lstm(50,50) \\
MountainCar & & 92.53 & lstm(50,50) \\
MountainCar & \checkmark & 88.2 & lstm(50,50) \\
Taxi-v3 & & -148.23 & $\text{gen}(0.2,0.2)$, $p_\text{unicx}$ \\
Taxi-v3 & \checkmark & -150.44 & human \\
BipedalWalker-v2 & & -0.15 & lstm(50,50)\\
\end{tabular}

However, comparing results for $T_\text{small}$ versus $T_\text{neural}$ proves that genetic developers have been intstrumental to the quality of these neural networks - this is to be expected with Priority Queue Training (see sec. \ref{sec:neural}).

\section{Discussion}

We have introduced a neurogenetic programming framework, demonstrated its efficacy and advantages over simpler program induction methods.

We believe that this framework can become a basis for many future methods - new methods of program synthesis can be built into the \emph{Instant Scrum} framework as developers and combined with existing ones as necessary.
In particular, one type of developer currently absent from our experiments is a \emph{neural mutation} - a neural network that modifies existing programs and can be trained to modify them in a way that improves their performance.
Another important direction is applying the framework to more specialized tasks like robotics or healthcare decision support. 

\footnotesize
\section*{Acknowledgements}

This work was funded by the European Union’s Horizon 2020 research and innovation programme under grant agreement 812882. This work is part of "Personal Health Interfaces Leveraging HUman-MAchine Natural interactionS" (PhilHumans) project: \url{https://www.philhumans.eu}

\normalsize

\end{multicols}

\begin{table}[]
    \centering
    \begin{tabular}{c|c|c|c|c|c|c|c}
         Environment & \multicolumn{2}{c}{CartPole-v1} & \multicolumn{2}{c}{MountainCarContinuous-v0} & \multicolumn{2}{c}{Taxi-v3} & BipedalWalker-v2 \\
         \midrule
         Initial programs & & 20.48 & & -6.55 & & -150.44 & \\
         \midrule
         $T_\text{small}$  &    60.93 &    143.91 &     \textbf{92.53} &     88.20 &   -148.23 &   -150.44 &     -0.16\\
         $T_\text{large}$ & \textbf{157.35} &     57.47 &     91.65 &     91.42 &    \textbf{-32.12} &   -150.44 &      \textbf{8.13} \\ 
         $T_\text{genetic}$& - & 59.12 & - & 0 & - & -47.54 & - \\ 
         $T_\text{neural}$ & 71.38 & 96.64 & 88.41 & 91.38 & -198.9 & -150.44 & 6.17 \\
         \midrule
         Leaderboard threshold & 195 & 195 & 90 & 90 & 0 & 0 & 300 \\ 
    \end{tabular}
    \caption{Averaged 100-episode reward acheived by the best program in each category}
    \label{tab:results}
\end{table}

\begin{multicols}{2}

\bibliography{bib}

\end{multicols}

\end{document}